\documentclass[10pt,twocolumn,letterpaper]{article}
\usepackage{cvpr}              
\usepackage[dvipsnames, table]{xcolor}

\usepackage{graphicx}
\usepackage{amsmath}
\usepackage{amssymb}
\usepackage{booktabs}

\usepackage{bm,bbm}
\usepackage{amsmath}
\usepackage{tabularx}
\usepackage{arydshln}
\usepackage{enumerate}
\usepackage[shortlabels]{enumitem}

\usepackage{adjustbox}
\usepackage{multirow}
\usepackage{graphicx}
\usepackage{makecell}

\usepackage[accsupp]{axessibility}
\newcommand{\paragraphTitle}[1]{\vspace{1mm}\noindent{\textbf{#1}\hspace{1mm}}}

\usepackage[pagebackref,breaklinks,colorlinks]{hyperref}

\usepackage[capitalize]{cleveref}
\crefname{section}{Sec.}{Secs.}
\Crefname{section}{Section}{Sections}
\Crefname{table}{Table}{Tables}
\crefname{table}{Tab.}{Tabs.}

\def\cvprPaperID{3052} 
\def\confName{CVPR}
\def\confYear{2023}

\def\eg{\textit{e.g.}}
\def\ie{\textit{i.e.}}

\def\etal{\textit{et al.}}

\definecolor{mygray}{gray}{0.9}

\newcommand{\YK}[1]{\textcolor{RedOrange}{{#1}}}

\newcommand{\Skip}[1]{}

\makeatletter
\def\thickhline{%
  \noalign{\ifnum0=`}\fi\hrule \@height \thickarrayrulewidth \futurelet
   \reserved@a\@xthickhline}
\def\@xthickhline{\ifx\reserved@a\thickhline
               \vskip\doublerulesep
               \vskip-\thickarrayrulewidth
             \fi
      \ifnum0=`{\fi}}
\makeatother
\newlength{\thickarrayrulewidth}
\begin{document}

\title{Feature Separation and Recalibration for Adversarial Robustness}


\author{
    \begin{tabular}[t]{cccc} Woo Jae Kim & Yoonki Cho & Junsik Jung & Sung-Eui Yoon \end{tabular}\vspace{1.5mm} \\
    Korea Advanced Institute of Science and Technology (KAIST) \\
}

\maketitle

\begin{abstract}

Deep neural networks are susceptible to adversarial attacks due to the accumulation of perturbations in the feature level, and numerous works have boosted model robustness by deactivating the non-robust feature activations that cause model mispredictions.
However, we claim that these malicious activations still contain discriminative cues and that with recalibration, they can capture additional useful information for correct model predictions.
To this end, we propose a novel, easy-to-plugin approach named \emph{Feature Separation and Recalibration (FSR)} that recalibrates the malicious, non-robust activations for more robust feature maps through Separation and Recalibration.
The Separation part disentangles the input feature map into the robust feature with activations that help the model make correct predictions and the non-robust feature with activations that are responsible for model mispredictions upon adversarial attack.
The Recalibration part then adjusts the non-robust activations to restore the potentially useful cues for model predictions.
Extensive experiments verify the superiority of \emph{FSR} compared to traditional deactivation techniques and demonstrate that it improves the robustness of existing adversarial training methods by up to 8.57\% with small computational overhead.
Codes are available at \url{https://github.com/wkim97/FSR}.

\Skip{
Deep neural networks are susceptible to adversarial attacks due to the accumulation of perturbations in the feature level, and numerous works have boosted model robustness by deactivating the non-robust feature activations that cause model mispredictions. However, we claim that these malicious activations still contain discriminative cues and that with recalibration, they can capture additional useful information for correct model predictions. To this end, we propose a novel, easy-to-plugin approach named Feature Separation and Recalibration (FSR) that recalibrates the malicious, non-robust activations for more robust feature maps through Separation and Recalibration. The Separation part disentangles the input feature map into the robust feature with activations that help the model make correct predictions and the non-robust feature with activations that are responsible for model mispredictions upon adversarial attack. The Recalibration part then adjusts the non-robust activations to restore the potentially useful cues for model predictions. Extensive experiments verify the superiority of FSR compared to traditional deactivation techniques and demonstrate that it improves the robustness of existing adversarial training methods by up to 8.57
}

\end{abstract}

\section{Introduction}
Despite the advancements of deep neural networks (DNNs) in computer vision tasks~\cite{resnet, yolo, vqa, pplr}, they are vulnerable to adversarial examples~\cite{lbfgs, fgsm} that are maliciously crafted to subvert the decisions of these models by adding imperceptible noise to natural images.
Adversarial examples are also known to be successful in real-world cases, including autonomous driving~\cite{driving} and biometrics~\cite{arcface, face}, and to be effective even when target models are unknown to the attacker~\cite{delving, lbfgs, ada}.
Thus, it has become crucial to devise effective defense strategies against this insecurity.

To this end, numerous defense techniques have been proposed, including defensive distillation~\cite{distillation}, input denoising~\cite{hgd}, and attack detection~\cite{robust-detection, squeezing}.
Among these methods, adversarial training~\cite{fgsm, pgd}, which robustifies a model by training it on a set of worst-case adversarial examples, has been considered to be the most successful and popular.

Even with adversarial training, however, small adversarial perturbations on the pixel-level accumulate to a much larger degree in the intermediate feature space and ruin the final output of the model~\cite{fd}.
To solve this problem, recent advanced methods disentangled and deactivated the non-robust feature activations that cause model mispredictions.
Xie \etal~\cite{fd} applied classical denoising techniques to deactivate disrupted activations, and Bai \etal~\cite{cas} and Yan \etal~\cite{cifs} deactivated channels that are irrelevant to correct model decisions.
These approaches, however, inevitably neglect discriminative cues that potentially lie in these non-robust activations.
Ilyas \etal~\cite{NotBugsFeatures} have shown that a model can learn discriminative information from non-robost features in the input space.
Based on this finding, we argue that there exist potential discriminative cues in the non-robust activations, and deactivating them could lead to loss of these useful information that can provide the model with better guidance for making correct predictions.

For the first time, we argue that with appropriate adjustment, the non-robust activations that lead to model mispredictions could recapture discriminative cues for correct model decisions.
To this end, we propose a novel Feature Separation and Recalibration (FSR) module that aims to improve the \textit{feature robustness}.
We first separate the intermediate feature map of a model into the malicious \textit{non-robust} activations that are responsible for model mispredictions and the \textit{robust} activations that still provide useful cues for correct model predictions even under adversarial attacks.
Exploiting only the robust feature just like the existing methods~\cite{fd, cas, cifs}, however, could lead to loss of potentially useful cues in the non-robust feature.
Thus, we recalibrate the non-robust activations to capture cues that provide additional useful guidance for correct model decisions. 
These additional cues can better guide the model to make correct predictions and thus boost its robustness.

\begin{figure}[t]
    \centering
    \includegraphics[width=1.0\columnwidth]{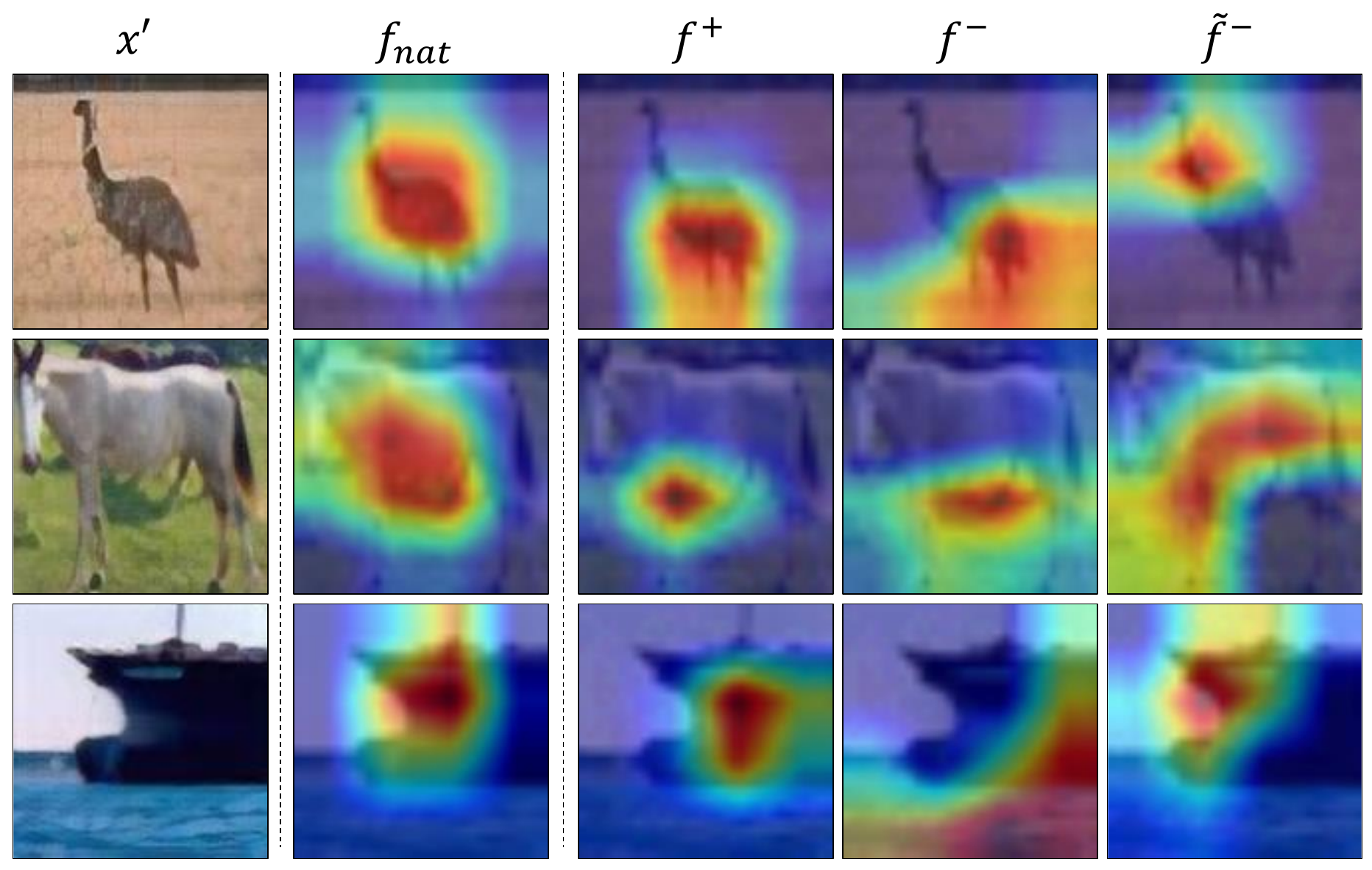}
    \caption{ 
        Visualization of attention maps~\cite{gradcam} on the features of natural images on a naturally trained model ($f_{nat}$) and the robust ($f^{+}$), non-robust ($f^{-}$), and recalibrated features ($\tilde{f}^{-}$) of adversarial examples ($x'$) obtained from an adversarial training~\cite{pgd} model equipped with our FSR module on CIFAR-10 dataset.
        The robust feature captures discriminative cues regarding the ground truth class, while the non-robust feature captures irrelevant cues.
        To further boost feature robustness, we recalibrate the non-robust feature ($f^{-} \rightarrow \tilde{f}^{-}$) and capture additional useful cues for model predictions.
        Adversarial images $x'$ are upscaled for clearer visual representations.
    }
    \label{fig:intro-gradcam}
    \vspace{-3mm}
\end{figure}


Fig.~\ref{fig:intro-gradcam} visualizes the attention maps~\cite{gradcam} on the features of natural images by a naturally trained model ($f_{nat}$) and the robust ($f^{+}$), non-robust ($f^{-}$), and recalibrated features ($\tilde{f}^{-}$) on adversarial examples ($x'$) obtained from an adversarial training~\cite{pgd} model equipped with our FSR module. 
Given an adversarial example, while the non-robust feature ($f^{-}$) captures cues irrelevant to the ground truth class, the robust feature ($f^{+}$) captures discriminative cues (\eg, horse's leg). 
Our FSR module recalibrates the non-robust activations ($f^{-} \rightarrow \tilde{f}^{-}$), which are otherwise neglected by the existing methods, and restores additional useful cues not captured by the robust activations (\eg, horse's body).
With these additional cues, FSR further boosts the model's ability to make correct decisions on adversarial examples.

Thanks to its simplicity, our FSR module can be easily plugged into any layer of a CNN model and is trained with the entire model in an end-to-end manner.
We extensively evaluate the robustness of our FSR module on benchmark datasets against various white-box and black-box attacks and demonstrate that our approach improves the robustness of different variants of adversarial training (Sec.~\ref{sec:exp-robustness}) with small computational overhead (Sec.~\ref{sec:exp-comp}).
We also show that our approach of recalibrating non-robust activations is superior to existing techniques~\cite{cas, fd, cifs} that simply deactivate them (Sec.~\ref{sec:exp-robustness}).
Finally, through ablation studies, we demonstrate that our Separation stage can effectively disentangle feature activations based on their effects on model decision and that our Recalibration stage successfully recaptures useful cues for model predictions (Sec.~\ref{sec:exp-ablation}).

In summary, our contributions are as follow:
\begin{itemize}
    \item In contrast to recent methods that deactivate distorted feature activations, we present a novel point of view that these activations can be recalibrated to capture useful cues for correct model decisions. 

    \item We introduce an easy-to-plugin \textit{Feature Separation and Recalibration (FSR)} module, which separates non-robust activations from feature maps and recalibrates these feature units for additional useful cues.

    \item Experimental results demonstrate the effectiveness of our FSR module on various white- and black-box attacks with small computational overhead and verify our motivation that recalibration restores discriminative cues in non-robust activations.
    
    \Skip{and verify our novel point of view.}
    \Skip{Extensive experiments show the superiority of our FSR module against existing feature deactivation methods and verify its effectiveness on improving the robustness of various adversarial training techniques under both white-box and black-box attacks.} 
    
\end{itemize}
\section{Related Works}

\begin{figure*}[t]
    \centering
    \includegraphics[width=0.8\textwidth]{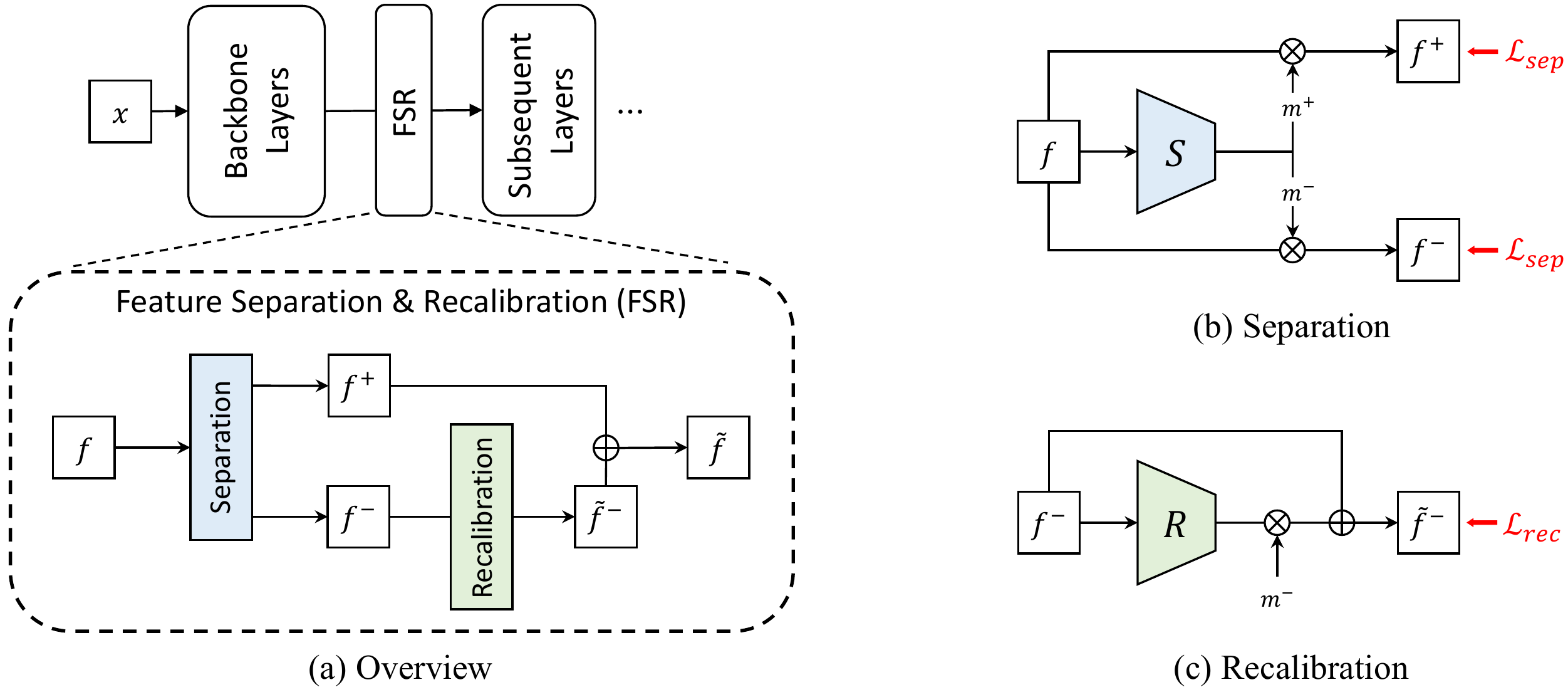}
    \vspace{-3mm}
    \caption{ 
        (a) Overview of the Feature Separation and Recalibration module.
        During the (b) Separation stage, we disentangle the input feature $f$ into the robust feature $f^{+}$ and the non-robust feature $f^{-}$ by applying the learnable positive mask $m^{+}$ and the negative mask $m^{-}$, respectively.
        Then, during the (c) Recalibration stage, we recalibrate the activations of $f^{-}$ into $\tilde{f}^{-}$ to restore the useful cues for correct model predictions.
        Finally, we combine the recalibrated feature and the robust feature to obtain the output feature $\tilde{f}$ and pass it down to subsequent layers of the model.
    }
    \vspace{-3mm}
    \label{fig:overview}
\end{figure*}

\subsection{Adversarial Training as Adversarial Defense}
Adversarial training guides a model to be robust against adversarial attacks by training it with adversarially generated data and has been widely considered as one of the most effective defense strategies.
It solves the following minimax optimization problem:
\begin{equation}
    \min_{\theta} \mathbb{E}_{(x, y) \sim \mathcal{D}} \left[ \max_{\delta} \mathcal{L}_{cls}(F_{\theta}(x + \delta), y) \right],
    \label{eq:1}
\end{equation}
\Skip{where $F$ is a DNN model with parameter $\theta$,}
where $F_{\theta}$ is a model parameterized by $\theta$,
$x$ is a natural image with label $y$ from dataset $\mathcal{D}$, $\delta$ is a perturbation bounded within the $\ell_p$-norm of magnitude $\epsilon$ such that $\left\| \delta \right\|_p \le \epsilon$, and $\mathcal{L}_{cls}(\cdot, \cdot)$ represents the classification loss. 
The inner maximization aims to find the strongest possible perturbation $\delta$ that maximizes the classification loss, and the outer minimization learns the model to minimize the loss with respect to the worst-case adversarial examples.
To optimize the inner maximization, Goodfellow \etal~\cite{fgsm} used the Fast Gradient Sign Method (FGSM), and Madry \etal~\cite{pgd} used the Projected Gradient Descent (PGD) attack.

Many variants of adversarial training have also been studied in recent years.
ALP~\cite{alp} reduced the distance between the logits from a natural image and its adversarial counterpart.
TRADES~\cite{trades} decomposed the prediction error on adversarial examples into the natural error and the boundary error to improve both robustness and accuracy.
MART~\cite{mart} additionally considered misclassified  examples during training.
Inspired by curriculum learning~\cite{curriculum}, CAT~\cite{cat} and FAT~\cite{fat} trained models with increasingly stronger adversarial examples to improve generalization.
SEAT~\cite{seat} proposed a self-ensemble method that combines the weights of different models through the training process, and S$^2$O~\cite{s2o} applied the second-order statistics to the model weights to improve adversarial training robustness.
Thanks to its simplicity, our method can be easily plugged into any of these adversarial training methods to further improve their robustness.

\subsection{Adversarial Defense on Feature Space}
In a parallel line of research, it has been found that some prior models learn non-robust features from the dataset~\cite{NotBugsFeatures} and that input perturbations of adversarial examples are often accumulated through intermediate layers to misguide the final prediction~\cite{fd}.
To solve these problems, several works tried to learn robust feature representations by modifying the network structure or applying regularizations.
Galloway \etal~\cite{bn-galloway}, Benz \etal~\cite{bn-benz}, and Wang \etal~\cite{bn-wang} studied the adversarial vulnerability from the perspective of batch normalization. 
Dhillon \etal~\cite{sap} applied pruning to a random set of activations, especially those with small magnitudes, and Madaan \etal~\cite{anp-vs} pruned out activations that are vulnerable to adversarial attacks.
Mustafa \etal~\cite{pcl} proposed a class-wise feature disentanglement and pushed the centers of each class from each other to learn more discriminative feature representations.
 \Skip{\YK{for $\sim$}.}

There also have been attempts to reduce abnormalities in the feature maps by explicitly manipulating the feature activations.
Xiao \etal~\cite{kWTA} proposed \textit{k}-Winner-Takes-All activation to deactivate all feature units except for \textit{k} units with the largest magnitudes.
Xu \etal~\cite{pse} interpreted the effects of adversarial perturbations on the pixel, image, and network levels and masked out feature units sensitive to perturbations.
Zoran \etal~\cite{attention} applied attention mechanism to emphasize important regions on the feature map.
Xie \etal~\cite{fd} proposed Feature Denoising (FD) that applies classical denoising techniques to deactivate abnormal activations.
Bai \etal~\cite{cas} and Yan \etal~\cite{cifs} studied the effects of perturbations on the feature activation from the channel perspective and proposed Channel Activation Suppression (CAS) and Channel-wise Importance-based Feature Selection (CIFS), respectively, to deactivate the activation of non-robust channels.

In contrast to these methods, we propose a recalibration strategy.
Existing deactivation strategies simply discard non-robust feature activations responsible for model mistakes.
Taking a step further, we adjust such activations to instead \textit{recapture} potentially discriminative cues and thus boost model robustness.

\section{Methods}
\label{sec:methods}
While the distorted feature activations upon adversarial attack are known to be responsible for model mispredictions, we argue that with adjustment, we can recapture useful cues for model predictions.
To fully utilize such potentially useful cues, we present a novel Feature Separation and Recalibration (FSR) module (Fig.~\ref{fig:overview}) that can restore these cues through a separation-and-recalibration scheme.
During the Separation stage, we disentangle the feature map into the \textit{robust} and \textit{non-robust} features by masking out the non-robust and robust activations, respectively.
Then, during the Recalibration stage, we recalibrate the activations of the non-robust features such that they provide useful information for correct model prediction.
Thanks to its simplicity, as shown in Fig.~\ref{fig:overview}, FSR module can be inserted to any layer of a model to improve its robustness.
We elaborate on the details of our FSR module in the following subsections.

\subsection{Feature Separation}
\label{ssec:3.1}
As compared to recent approaches \cite{fd, cas, cifs} that robustify feature maps by deactivating non-robust activations, we aim to recalibrate them to restore potentially useful cues for model predictions.
To this end, during the Separation stage, we extract the non-robust activations to be recalibrated from the input intermediate feature map. 
To determine which activation is robust or non-robust, we introduce the Separation Net $S$, which learns the robustness of each feature unit.
Given the intermediate feature map $f \in \mathbb{R}^{C \times H \times W}$ as input, where $C$, $H$, and $W$ represent the channel, height, and width dimensions of $f$, respectively, the Separation Net outputs a robustness map $r \in \mathbb{R}^{C \times H \times W}$ that represents a robustness scores of the corresponding units of $f$, where a higher score means a more robust feature activation.

In order to extract the non-robust feature, we disentangle the feature map into the robust feature and the non-robust feature in an element-wise manner based on the robustness score.
One way to achieve this goal is to apply to the feature map a binary mask $b\in \{0, 1\}^{C \times H \times W}$ generated based on the robustness score.
However, such discrete sampling is non-differentiable~\cite{gumbel} and discontinuous~\cite{soft-topk}, which could cause gradient masking that would give a false sense of robustness~\cite{practical, obfuscated}.

To avoid such problem, we approximate a binary mask $b$ with a differentiable soft mask $m \in [0, 1]^{C \times H \times W}$ using Gumbel softmax~\cite{gumbel} such that:
\begin{equation}
        m = \frac{e^{((\log{(\sigma(r))} + g_1) / \tau)}}
        {e^{((\log{(\sigma(r))} + g_1) / \tau)} + e^{((\log{(1 - \sigma(r))} + g_2) / \tau)}},
    \label{eq:mask}
\end{equation}
where $r$ is the robustness map, and $\sigma$ is a sigmoid function used to normalize the robustness map.
$g_1$ and $g_2$ represent the samples drawn from the Gumbel distribution such that $g = -\log(-\log(u))$, where $u \sim \text{Uniform}(0, 1)$, and $\tau$ is a temperature used to control the effects of $g_1$ and $g_2$.
Note that during inference, to avoid stochasticity from sampling $g_1$ and $g_2$, we fix them as $-\log(-\log(u_c))$, where $u_c \in \mathbb{R}^{C \times H \times W}$ is the expected value of Uniform distribution.
By computing two-class Gumbel softmax between the normalized robustness map and its inverted version, we obtain a mask $m$ with values close to 1 for high robustness scores and values close to 0 for low robustness scores.
Then, to mask out the non-robust activations from the input feature $f$ and obtain the robust feature $f^{+}$, we compute element-wise product between the feature $f$ and the positive mask $m^{+} = m$ such that $f^{+} = m^{+} \otimes f$.
\Skip{Similarly, we can mask out the robust activations and obtain the non-robust feature $f^{-}$ by computing the element-wise product between the feature $f$ and the negative mask $m^{-} = 1 - m$ such that $f^{-} = m^{-} \otimes f$.}
Similarly, we obtain non-robust feature $f^{-} = m^{-} \otimes f$ by masking out the robust activations with the negative mask $m^{-} = 1 - m$.


Without any guidance, however, the Separation Net may not learn the correct robustness score for each activation.
To this end, we design an objective that guides the Separation Net to learn robustness scores specifically based on the influence of feature activations on model making a correct or incorrect prediction.
We attach an MLP-based auxiliary layer $h$ that takes each of the two feature $f^{+}$ and $f^{-}$ as inputs and outputs prediction scores $p^{+}$ and $p^{-}$, respectively.
Then, we compute the separation loss $\mathcal{L}_{sep}$ as follows:
\begin{equation}
    \mathcal{L}_{sep} = - \sum^{N}_{i=1} (y_i \cdot \log(p^{+}_i) + y'_i \cdot \log(p^{-}_i)),
    \label{eq:l_sep}
\end{equation}
where $N$ is the number of classes, $y$ is the ground truth label, and $y'$ is the label corresponding to the wrong class with the highest prediction score from the final model output.
By training the auxiliary layer to make correct predictions based on the activations that are preserved from the positive mask $m^{+}$, $\mathcal{L}_{sep}$ guides the Separation Net to assign high robustness scores to units that help the auxiliary layer make correct predictions.
At the same time, we aim to disentangle the highly disrupted activations that cause the model to make mispredictions upon adversarial attack.
$\mathcal{L}_{sep}$ also guides the Separation Net to assign low robustness scores to non-robust activations that are specifically responsible for the most probable misprediction.

With $\mathcal{L}_{sep}$, the Separation Net can effectively separate the robust and the non-robust feature activations.
Discarding the non-robust activations is one way to improve feature robustness; however, this approach would ignore potentially useful cues that can be recaptured through recalibration (see Sec.~\ref{sec:exp-ablation}).
In the following subsection, we discuss how we recalibrate the disentangled non-robust activations to capture additional useful cues for improved feature robustness.

\subsection{Feature Recalibration}
Exploiting only the robust feature obtained through the Separation stage just like recent techniques~\cite{fd, cas, cifs} could lead to loss of potentially useful cues for model predictions that could further boost model robustness.
Therefore, for the first time, we adjust the non-robust feature activations to capture the additional useful cues during the Recalibration stage.
We first introduce the Recalibration Net $R$ that takes the non-robust feature $f^{-}$ as input and outputs recalibrating units that are designed to adjust the activations of $f^{-}$ accordingly.
To recalibrate the non-robust activations designated by the robustness map, we apply the negative mask $m^{-}$ to the recalibrating units.
Finally, we compute the recalibrated feature $\tilde{f}^{-}$ by adding the result to $f^{-}$, \ie, 
$\tilde{f}^{-} = f^{-} + m^{-} \otimes R(f^{-})$.

The goal of the Recalibration stage is to make the non-robust activations recapture cues that can help the model make correct decisions.
To guide the Recalibration Net to achieve this goal, we again attach the auxiliary layer $h$ after the recalibrated feature $\tilde{f}^{-}$ and compute the recalibration loss $\mathcal{L}_{rec}$ as follows:
\begin{equation}
    \mathcal{L}_{rec} = - \sum^{N}_{i=1} y_i \cdot \log(\tilde{p}^{-}_i),
    \label{eq:l_rec}
\end{equation}
where $\tilde{p}^{-}$ is the output prediction score of the auxiliary layer given $\tilde{f}^{-}$ as input.
By training the same auxiliary layer to make correct decisions based on the recalibrated feature, we guide the Recalibration Net to adjust the non-robust activations such that they provide cues relevant to the ground truth class.
After the Recalibration stage, we add the robust feature $f^{+}$ and the recalibrated non-robust feature $\tilde{f}^{-}$ in an element-wise manner to obtain the output feature map $\tilde{f} = f^{+} + \tilde{f}^{-}$, which is passed to subsequent layers of the model.
Through the Recalibration stage, we can capture additional useful cues from the non-robust activations, which are neglected in previous approaches.

\begin{table*}[t]
	\begin{center}
	    \resizebox{0.95\linewidth}{!}
		{\begin{tabular}{c|c c c c c c|c c c c c c}
        \hline
        \textbf{\textit{ResNet-18}} & \multicolumn{6}{c|}{CIFAR-10} & \multicolumn{6}{c}{SVHN} \\ \hline 
        Method & Natural & FGSM & PGD-20 & PGD-100 & C\&W & Ensemble & Natural & FGSM & PGD-20 & PGD-100 & C\&W & Ensemble \\
        \hline \hline
        {AT}            & \textbf{85.02} & 56.21 & 48.22 & 46.37 & 47.38 & 45.51         
                        & 91.21 & 55.55 & 40.85 & 37.54 & 40.61 & 37.41 \\
        {AT + FSR}      & 81.46 & \textbf{58.07} & \textbf{52.47} & \textbf{51.02} & \textbf{49.44} & \textbf{48.34}    
                        & \textbf{91.28} & \textbf{60.46} & \textbf{43.94} & \textbf{39.01} & \textbf{43.22} & \textbf{38.81} \\
        \hline
        {TRADES}        & \textbf{86.31} & 57.21 & 50.74 & 49.44 & 48.66 & 47.89         
                        & 90.99 & 61.31 & 47.12 & 43.55 & 45.48 & 42.99 \\
        {TRADES + FSR}  & 84.49 & \textbf{58.29} & \textbf{52.27} & \textbf{51.28} & \textbf{49.92} & \textbf{49.28}         
                        & \textbf{91.39} & \textbf{68.85} & \textbf{51.49} & \textbf{47.50} & \textbf{46.70} & \textbf{46.17} \\
        \hline
        {MART}          & 82.73 & 56.65 & 50.88 & 49.15 & 47.21 & 45.98         
                        & \textbf{90.50} & 58.21 & 43.61 & 40.43 & 42.20 & 40.07 \\
        {MART + FSR}    & \textbf{83.28} & \textbf{59.55} & \textbf{54.80} & \textbf{53.69} & \textbf{48.98} & \textbf{48.36}
                        & 89.87 & \textbf{61.06} & \textbf{46.51} & \textbf{42.94} & \textbf{43.89} & \textbf{42.40} \\
                        
        \hline
        \end{tabular}}
	\end{center}
        \vspace{-3mm}
	\caption{
	    Robustness (accuracy (\%)) of adversarial training strategies (AT, TRADES, MART) with (+ FSR) and without our FSR module against diverse white-box attacks on ResNet-18. Better results are marked in \textbf{bold}.
	}
	\label{table:robustness-resnet}
\end{table*}

\begin{table*}[t]
	\begin{center}
	    \resizebox{0.95\linewidth}{!}
		{\begin{tabular}{c|c c c c c c|c c c c c c}
        \hline
         \textbf{\textit{VGG16}} & \multicolumn{6}{c|}{CIFAR-10} & \multicolumn{6}{c}{SVHN} \\ \hline
        Method & Natural & FGSM & PGD-20 & PGD-100 & C\&W & Ensemble & Natural & FGSM & PGD-20 & PGD-100 & C\&W & Ensemble \\
        \hline
        \hline
        AT              & \textbf{80.56} & 53.47 & 47.17 & 45.58 & 45.82 & 43.71         
                        & 89.59 & 54.88 & 40.27 & 36.90 & 39.46 & 36.62 \\
        AT + FSR        & 80.06 & \textbf{54.40} & \textbf{49.82} & \textbf{48.82} & \textbf{47.28} & \textbf{46.24}         
                        & \textbf{91.44} & \textbf{65.01} & \textbf{45.99} & \textbf{39.07} & \textbf{43.08} & \textbf{38.15} \\
        \hline
        TRADES          & \textbf{82.44} & 53.92 & 47.39 & 46.20 & 44.80 & 44.20         
                        & 90.48 & 61.50 & 45.99 & 40.00 & 42.82 & 39.27 \\
        TRADES + FSR    & 80.78 & \textbf{55.48} & \textbf{49.95} & \textbf{49.03} & \textbf{46.28} & \textbf{45.90}         
                        & \textbf{91.89} & \textbf{69.25} & \textbf{54.56} & \textbf{47.81} & \textbf{46.66} & \textbf{44.10} \\
        \hline
        MART            & 76.11 & 54.86 & 51.06 & 50.16 & 43.53 & 43.01     
                        & 89.95 & 59.03 & 42.89 & 38.73 & 39.12 & 37.64 \\
        MART + FSR      & \textbf{79.18} & \textbf{56.41} & \textbf{52.69} & \textbf{52.13} & \textbf{44.49} & \textbf{44.20}         
                        & \textbf{90.60} & \textbf{62.28} & \textbf{47.17} & \textbf{42.50} & \textbf{43.44} & \textbf{40.73} \\
        \hline
        \end{tabular}}
	\end{center}
        \vspace{-3mm}
	\caption{
	    Robustness (accuracy (\%)) of adversarial training strategies (AT, TRADES, MART) with (+ FSR) and without our FSR module against diverse white-box attacks on VGG16. Better results are marked in \textbf{bold}.
	}
        \vspace{-3mm}
	\label{table:robustness-vgg}
\end{table*}

\subsection{Model Training}
The proposed FSR module can be easily inserted to any layer of a model and is trained with the entire model in an end-to-end manner thanks to its simplicity.
We can also apply the proposed method with any classification loss $\mathcal{L}_{cls}$ for different types of adversarial training~\cite{pgd, trades, mart}, and the overall objective function is as follows:
\begin{equation}
    \mathcal{L} = \mathcal{L}_{cls} + 
    \frac{1}{\vert L \vert}\sum_{l \in L} \left( {\lambda_{sep}} \cdot \mathcal{L}^l_{sep} + 
                          {\lambda_{rec}} \cdot \mathcal{L}^l_{rec} \right),
\end{equation}
where $L$ represents the set of positions in which the FSR module is inserted, and $\mathcal{L}^l_{sep}$ and $\mathcal{L}^l_{rec}$ each represents the separation loss and the recalibration loss applied on the FSR module at $l$-th layer.
The hyperparameters $\lambda_{sep}$ and $\lambda_{rec}$ are used to control the weights of $\mathcal{L}_{sep}$ and $\mathcal{L}_{rec}$, respectively.
Addition of this simple FSR module can improve the robustness of adversarial training methods against both white-box and black-box attacks with small computational overhead, as described in the following section. 

\section{Experiments}
\label{sec:exp}

\subsection{Experimental Setups}
\vspace{0.5ex}\noindent
\textbf{Evaluation Protocols.}
We evaluate our method on CIFAR-10/100~\cite{cifar10}, SVHN~\cite{svhn}, and Tiny ImageNet~\cite{learnable} datasets using ResNet-18~\cite{resnet}, VGG16~\cite{vgg}, and WideResNet-34-10~\cite{wideresnet} as the baseline models.
We apply our method to PGD adversarial training (AT)~\cite{pgd} and other variants of adversarial training, \ie, TRADES~\cite{trades} and MART~\cite{mart}, to verify its wide applicability.
For training, we use PGD-10~\cite{pgd} with perturbation bound $\epsilon$ = 8/255 (step size $\epsilon$/4 for CIFAR-10/100 and Tiny ImageNet, $\epsilon$/8 for SVHN) under $\ell_\infty$-norm to craft adversarial examples.
For evaluation, we use FGSM~\cite{fgsm}, PGD-20~\cite{pgd} (step size $\epsilon$/10), PGD-100~\cite{pgd} (step size $\epsilon$/10), and C\&W~\cite{c&w} (PGD optimization for 30 steps with step size $\epsilon$/10) bounded within $\epsilon$ = 8/255 under $\ell_\infty$-norm.
As suggested by Carlini \etal~\cite{evaluating}, to better compare the robustness of different defense techniques, we also report the average per-example Ensemble robustness of the model as formulated below~\cite{adt}:
\begin{equation}
    \text{Ensemble} = \frac{1}{N_{test}}\sum^{N_{test}}_{i=1}\min_{a \in \mathcal{A}} \mathbbm{1}(F_\theta(a(x)) = y),
\end{equation}
where $N_{test}$ is the number of images in the test dataset, $\mathbbm{1}(\cdot)$ is an indicating function, $F_\theta$ is a target model with parameter $\theta$, $y$ is the ground truth label, and $\mathcal{A}$ is a set of adversarial attacks (FGSM, PGD-20, PGD-100, and C\&W).

\vspace{0.5ex}\noindent
\textbf{Implementation Details.}
We train all models for 100 epochs using an SGD optimizer (momentum 0.9, weight decay $5\times10^{-4}$).
We set the initial learning rate to 0.1 for CIFAR-10/100 and Tiny ImageNet and to 0.01 for SVHN, and reduce it by a factor of 10 after 75-th and 90-th epochs.
We empirically set $\lambda_{sep} = 1$, $\lambda_{rec} = 1$, and $\tau$ for Gumbel softmax as $0.1$.
We implement the Separation Net as a series of three blocks each comprised of a convolutional layer, a batch normalization layer, and a ReLU activation, except for the last block which consists of a single convolutional layer.
The Recalibration Net also consists of a series of three blocks each comprised of a convolutional layer, a batch normalization layer, and a ReLU activation.
We insert our FSR module after \texttt{block4} for ResNet-18, \texttt{block4} for VGG16, and \texttt{block3} for WideResNet-34-10.

\subsection{Robustness Evaluation}
\label{sec:exp-robustness}

\vspace{0.5ex}\noindent
\textbf{Defense against White-box Attacks.}
To evaluate the ability of our FSR module to improve the model robustness of various adversarial training techniques, we report in Table~\ref{table:robustness-resnet} the effectiveness of applying FSR to three different methods (AT, TRADES, and MART) on ResNet-18.
Applying our FSR module consistently improves the robustness of all defense techniques under all of the individual attacks and the Ensemble attack.
Similar trends are observed in the SVHN dataset (Table~\ref{table:robustness-resnet}) and on VGG16 (Table~\ref{table:robustness-vgg}); for example, FSR improves the robustness of TRADES on VGG16 under PGD-20 by 8.57\% on the SVHN dataset.
By recalibrating the malicious non-robust activations, our method provides the model with additional useful cues for improved robustness.
With only a slight increase in the number of computations (see Sec.~\ref{sec:exp-comp}), we can improve the robustness of various adversarial training methods regardless of the dataset and the model.
Results on WideResNet-34-10, CIFAR-100, Tiny ImageNet are provided in the supplementary materials.

One observation is that our FSR module occasionally drops the accuracy on natural images.
This is because our method is designed to disentangle and recalibrate the ``malicious" activations that are \textit{intentionally} crafted to fool model predictions.
In natural images, since there are no intentionally disrupted feature activations, trying to identify and recalibrate these malicious cues can lead to potential loss of discriminative information and occasional accuracy drop.
Nevertheless, the natural accuracy drops only by a small amount or even improves upon the addition of FSR module (\eg, on SVHN), showing that this phenomenon is also dependent on the dataset.

\begin{table}[t]
	\begin{center}
	    \resizebox{\linewidth}{!}
		{\begin{tabular}{c|c c c c}
        \hline
        Method & TI-FGSM & DI-FGSM & $\mathcal{N}$Attack & AutoAttack \\
        \hline
        \hline
        AT              & 59.03 & 46.56 & 39.55 & 44.11 \\
        AT + FSR        & \textbf{62.53} & \textbf{50.50} & \textbf{52.65} & \textbf{46.41} \\
        \hline
        TRADES          & 59.95 & 49.23 & 43.45 & 46.81 \\
        TRADES + FSR    & \textbf{61.45} & \textbf{50.54} & \textbf{50.43} & \textbf{48.45} \\
        \hline
        MART            & 59.73 & 48.38 & 44.68 & 44.27 \\
        MART + FSR      & \textbf{62.68} & \textbf{51.42} & \textbf{53.76} & \textbf{46.55} \\
        \hline
        \end{tabular}}
	\end{center}
        \vspace{-3mm}
	\caption{
     Robustness (accuracy (\%)) of adversarial training strategies (AT, TRADES, MART) with (+ FSR) and without our FSR module against diverse black-box attacks and AutoAttack on CIFAR-10 using ResNet-18. Better results are marked in \textbf{bold}.
	}
	\label{table:black-box}
\end{table}

\vspace{0.5ex}\noindent
\textbf{Defense against Black-box Attacks and AutoAttack.}
To show that FSR improves adversarial training methods even under different types of attacks, we evaluate our method against a variety of black-box attacks and AutoAttack~\cite{autoattack}.
For black-box attacks, we use two transfer-based attacks -- TI-FGSM~\cite{tifgsm} and DI-FGSM~\cite{difgsm} -- crafted on a naturally trained ResNet-50, and $\mathcal{N}$Attack~\cite{nattack}, which is a strong query-based attack.
Following CAS~\cite{cas}, to evaluate each method against $\mathcal{N}$Attack, we sample 1,000 images from CIFAR-10 test set and limit the number of queries to 40,000.
AutoAttack, which is an ensemble of two Auto-PGD attacks~\cite{autoattack}, Fast Adaptive Boundary attack~\cite{fab}, Square attack~\cite{square}, has been shown to evaluate the robustness of defense techniques more reliably.
As shown in Table~\ref{table:black-box}, our method improves the robustness against diverse black-box attacks, especially against the stronger $\mathcal{N}$Attack.
Our method also improves the robustness against AutoAttack, showing that the improvement upon the addition of FSR module is truly thanks to making the model robust instead of obfuscated gradients~\cite{obfuscated} or improper evaluation.

\begin{table*}
\begin{center}
\resizebox{0.65\textwidth}{!}
{
\begin{tabular}{c|c|cccc|c|c}
\hline
Method & Natural & FGSM           & PGD-20         & PGD-100        & C\&W  & \cellcolor{mygray}Ensemble & \cellcolor{mygray}AutoAttack \\ \hline \hline
AT   & 85.02          & 56.21 & 48.22 & 46.37 & 47.38          & \cellcolor{mygray}45.51 & \cellcolor{mygray}44.11 \\ 
FD   & 85.14          & 56.81 & 48.54 & 46.70 & 47.72          & \cellcolor{mygray}45.82 & \cellcolor{mygray}44.57 \\
CAS  & \textbf{85.78} & 55.57 & 50.42 & 49.91 & \textbf{53.47} & \cellcolor{mygray}46.46 & \cellcolor{mygray}44.23 \\
CIFS & 79.87          & 56.53 & 49.80 & 48.17 & 49.89          & \cellcolor{mygray}47.26 & \cellcolor{mygray}43.94 \\ \hline
FSR (Ours) & 81.46    & \textbf{58.07} & \textbf{52.47} & \textbf{51.02} & 49.44 & \cellcolor{lightgray}\textbf{48.34}   & \cellcolor{lightgray}\textbf{46.41}      \\ \hline
\end{tabular}
}
\end{center}
\vspace{-3mm}
\caption{Comparison of robustness (accuracy (\%)) between existing methods and our method. All models are trained using AT with ResNet-18 on CIFAR-10. The best results are marked in \textbf{bold}, and more comprehensive Ensemble and AutoAttack are highlighted in grey.}
\label{table:existing-methods}
\end{table*}

\vspace{0.5ex}\noindent
\textbf{Comparison with Existing Methods.}
To verify the effectiveness of recalibrating non-robust activations, we report in Table~\ref{table:existing-methods} the comparison of our method to existing feature manipulation methods (FD, CAS, and CIFS).
We leave out from evaluation kWTA~\cite{kWTA} and SAP~\cite{sap}, which also manipulate feature activations for robustness, because they are known to cause gradient masking~\cite{adaptive, obfuscated}.
We report the robustness against various white-box attacks, the Ensemble of these attacks, and AutoAttack.
Our FSR outperforms the three methods under most white-box attacks.
CAS and CIFS excel at defending against C\&W because they exploit the weights of auxiliary classifiers or the gradients of their logit outputs to manipulate feature activations and thus enlarge the prediction margins on the feature space~\cite{cas}.
Our method does not adopt such technique and slightly lags behind CAS and CIFS under C\&W attack.
Nevertheless, our method achieves the highest Ensemble robustness, verifying that overall, our method is the most robust.
FSR also outperforms all methods under AutoAttack, showing that it is more reliable even under an ensemble of various white-box and black-box attacks.
As opposed to these approaches that deactivate the non-robust feature activations, our method instead recalibrates them to capture additional useful cues that help the model make correct predictions, thus improving the model robustness.

\subsection{Ablation Studies}
\label{sec:exp-ablation}
We first evaluate whether the individual components of FSR module work as desired by measuring the following: (a) classification accuracy of model and (b) weighted $k$-NN accuracy.
To evaluate (a) the classification accuracy of model, we pass down different features from FSR into the subsequent layers instead of the final feature $\tilde{f}$.
By doing so, we compare the \textit{model robustness} brought by these different features.

Different from the model robustness, we also explicitly measure the \textit{feature robustness} based on how well each feature captures discriminative cues corresponding to the ground truth class. 
To do so, we embed each feature among the features of natural images and measure (b) the weighted $k$-NN accuracy~\cite{unsup1, unsup2}.
For each arbitrary feature $f$, we first compute the weight $w_i$ for each neighbor $\bm{v_i}$ as its cosine similarity to $f$ with the temperature parameter $\gamma$ such that $w_i = \exp(\cos(f, \bm{v_i}) / \gamma)$.
Then, we compute the prediction score $s_c$ for each class $c$ by weighting the vote of each of the $k$-nearest neighbors $\mathcal{N}_k$ as $s_c = \sum_{i \in \mathcal{N}_k} w_i \cdot \mathbbm{1}(c_i = c)$.
Through this way, we measure robustness of each feature by determining how close it lies to the unperturbed features.

\begin{table}
	\begin{center}
	    \resizebox{0.9\columnwidth}{!}
		{\begin{tabular}{c|c c|c c}
                \hline
                \multicolumn{1}{c|}{\multirow{2}{*}{Method}} & \multicolumn{2}{c|}{(a) Classification} & \multicolumn{2}{c}{(b) Weighted $k$-NN} \\
                \cline{2-5}
                ~ &  Ensemble & AutoAttack & \enspace 5-NN \enspace  & \enspace 20-NN \enspace \\
                \hline
                \hline
                $f^{+}$ & 47.89 & 45.82 & 66.21 & 61.58 \\
                $f^{-}$ & 33.11 & 28.39 & 54.69 & 53.89 \\
                $\tilde{f}^{-}$ & 46.93 & 44.52 & 66.34 & 65.64 \\
                $\tilde{f}$ (Ours) & 48.34 & 46.41 & 70.91 & 65.88 \\ 
                \hline
                \end{tabular}}
        \end{center}
        \vspace{-3mm}
        	\caption{
                Ablation studies on the robustness of various feature maps obtained throughout our framework on CIFAR-10 using ResNet-18.
        	(a) Robustness (\%) of the model upon replacing the final feature map $\tilde{f}$ with different feature maps.
                (b) Top-1 accuracy (\%) of weighted $k$-NN on different feature maps. 
                }
        \vspace{-3mm}
	\label{table:abl-features}
\end{table}


\vspace{0.5ex}\noindent
\textbf{Evaluation on Separation.}
We show that our Separation stage does appropriately disentangle the non-robust feature from the input feature, which is an essential step in our method before recalibrating non-robust activations.
To do so, we evaluate how well the robust feature $f^{+}$ and the non-robust feature $f^{-}$ capture useful cues for correct model predictions.
We measure the classification accuracy of the model against the Ensemble (of FGSM, PGD-20, PGD-100, and C\&W) and AutoAttack upon using each feature (\ie, $\tilde{f} = f^{+}$ or $\tilde{f} = f^{-}$), whose results are shown in the left side (a) of Table~\ref{table:abl-features}.
We can observe that $f^{+}$ leads to much higher robustness against both Ensemble and AutoAttack than $f^{-}$, which drastically decreases the adversarial robustness.
This implies that while $f^{+}$ captures robust cues that lead to correct model decisions on adversarial examples, $f^{-}$ captures non-robust cues that are responsible for mispredictions upon attack, showing that our Separation stage well disentangles the input feature activations as intended.


On the right side (b) of Table~\ref{table:abl-features}, we report the 5-NN and 20-NN accuracies on $f^{+}$ and $f^{-}$.
Both cases show that $f^{+}$ results in higher accuracies than $f^{-}$.
This is because we disentangle $f^{+}$ as activations that help the model make correct predictions, and it thus captures more similar representations as the features of natural images than $f^{-}$ does.


\vspace{0.5ex}\noindent
\textbf{Evaluation on Recalibration.}
We also evaluate the ability of Recalibration to adjust the non-robust activations such that they capture cues that help the model make correct predictions.
We again test the classification accuracy of our model upon replacing the final feature map $\tilde{f}$ with the non-robust feature $f^{-}$ or the recalibrated non-robust feature $\tilde{f}^{-}$.
As shown in the left side (a) of Table~\ref{table:abl-features}, $\tilde{f}^{-}$ leads to huge improvements in robustness compared to $f^{-}$, showing that our Recalibration stage appropriately adjusts the non-robust activations to capture cues that help the model make correct predictions.

We can also observe that $f^{+}$, which is equivalent to simply suppressing the non-robust activations with our mask $m^{+}$, lags behind using both the robust and recalibrated features, \ie, $\tilde{f}=f^{+}+\tilde{f}^{-}$ (Ours).
This shows that recalibrating the non-robust activations restores additional useful cues for model decisions that are not captured by the robust activations and further improves the adversarial robustness, advocating for the necessity of our Recalibration stage.

Similar trends can also be observed on the $k$-NN accuracy as shown in the right side (b) of Table~\ref{table:abl-features}.
The recalibrated non-robust feature $\tilde{f}^{-}$ leads to significantly higher $k$-NN accuracy than $f^{-}$, and the accuracy of $f^{+}$ lags behind that of $\tilde{f} = f^{+} + \tilde{f}^{-}$ (Ours), again verifying the effectivness of our Recalibration stage.

\begin{table}
    \begin{center}
    \resizebox{\columnwidth}{!}
    {
        \begin{tabular}{c|c c c c c c}
        \hline
        Method     & FGSM & PGD-20 & PGD-100 & C\&W & Ensemble & AutoAttack \\ 
        \hline 
        \hline
        AT          & 56.21 & 48.22 & 46.37 & 47.38 & 45.51 & 44.11 \\ 
        \hline
        + FSR       & \textbf{58.07} & \textbf{52.47} & \textbf{51.02} & \textbf{49.44} & \textbf{48.34} & \textbf{46.41} \\
        w/o Sep     & 57.51 & 50.71 & 48.98 & 49.32 & 47.60 & 45.47 \\
        w/o Rec     & 57.67 & 50.06 & 48.54 & 49.41 & 47.32 & 44.96 \\
        \hline
        \end{tabular}
        }
        \end{center}
        \vspace{-3mm}
        \caption{
        Comparison of robustness (\%) of FSR applied on AT upon removing the Separation or the Recalibration stage. 
        Model and dataset used are ResNet-18 and CIFAR-10, respectively.
        Best results are marked in \textbf{bold}.
        }
        \vspace{-3mm}
    \label{table:abl-sep}
\end{table}

\vspace{0.5ex}\noindent
\textbf{Effectiveness of Separation and Recalibration.}
We verify the necessity of the Separation and the Recalibration stages by comparing the model robustness upon removing each of them from FSR.
As shown in Table~\ref{table:abl-sep}, removing the Separation stage (\ie, recalibrating the entire input feature map) leads to drop in adversarial robustness.
This is because recalibrating the entire feature map that contains both robust and non-robust activations leads to a suboptimal training of the Recalibration Net.
More specifically, the Recalibration Net learns to minimize the recalibration loss through the discriminative information that is already captured by the robust activations instead of recalibrating the non-robust activations.
Nevertheless, FSR without Separation still improves the robustness of AT, showing that Recalibration effectively captures useful cues for model predictions.

We also compare robustness as we remove the Recalibration stage and pass down only the robust feature into subsequent layers.
As shown in Table~\ref{table:abl-sep}, removing the Recalibration stage also decreases robustness.
This demonstrates the necessity of the Recalibration stage and also confirms that the non-robust activations contain additional useful cues which further boost model robustness.
Still, FSR without Recalibration improves the robustness compared to vanilla AT.
This shows that our Separation stage well disentangles the intermediate feature map based on feature robustness and outputs robust activations that provide useful cues for model predictions.

In the supplementary materials, we report additional ablation studies including the effects of different hyperparameters ($\lambda_{sep}$, $\lambda_{rec}$, $\tau$), effectiveness of Gumbel softmax, addition of FSR on different layers, choice of $y'$ in Eq.~\ref{eq:l_sep}, and effects of replacing robustness map $m$ with other strategies.

\subsection{Computational Efficiency}
\label{sec:exp-comp}
In Table~\ref{table:computation}, we provide the computational analysis of our method compared to a vanilla model in terms of the number of parameters (\# params (M)) and the number of floating point operations (FLOPs (G)).
With only a slightly more number of computations, we can improve the robustness of adversarial training and its variants with a significant margin.
Additionally, on the CIFAR-10 dataset, one training epoch of PGD-10 adversarial training on the vanilla ResNet-18 takes 114 seconds, while it takes 120 seconds on ResNet-18 with FSR module.
One epoch of evaluation on PGD-20 adversarial examples takes 19 seconds for the vanilla ResNet-18, while it takes 20 seconds for ResNet-18 equipped with FSR module.
With slight computational overhead, our FSR module improves the robustness of traditional adversarial training models.

\begin{table}
	\begin{center}
	\resizebox{\columnwidth}{!}
	{\begin{tabular}{c|c c|c c}
            \hline
            \multicolumn{1}{c|}{\multirow{2}{*}{Method}} & \multicolumn{2}{c|}{\textit{\textbf{VGG16}}} & \multicolumn{2}{c}{\textit{\textbf{ResNet-18}}}\\
            \cline{2-5}
             ~ & \# Params (M) & FLOPs (G) & \# Params (M) & FLOPs (G) \\
            \hline
            \hline
            Vanilla     & 15.25         & 0.6299    & 11.17         & 1.1133    \\
            + FSR       & 16.52         & 0.6701    & 12.43         & 1.1535    \\
            \hline
        \end{tabular}}
	\end{center}
        \vspace{-3mm}
	\caption{
	    Comparison of computational costs (\# params and FLOPs) on a vanilla model and a model with our FSR module.
	}
        \vspace{-3mm}
	\label{table:computation}
\end{table}



\section{Discussion}
Since the goal of our work is to improve the robustness of adversarial training models against adversarial examples, it lies on the assumption that the input images contain malicious perturbations designed to fool the model.
Thus, we deliberately design the Separation stage to disentangle activations that specifically lead the model to mispredictions.
However, natural images may not contain such malicious cues, and thus, our FSR module occasionally decreases the natural accuracy by a small amount.
A potential direction for future work could be applying curriculum learning~\cite{curriculum} to make FSR be better aware of the relationship between feature robustness and attack strength.

Adversarial attack poses a huge threat to the deployment of deep neural networks (DNN) in real-world applications~\cite{driving, arcface, face}.
In this regard, our work contributes positively to the research field by designing an easy-to-plugin module to robustify the DNN models against adversarial attacks.
Thanks to its simplicity, we expect that our method could also serve as a basis to design more robust DNN models in diverse and real-world applications.

\section{Conclusion}
In this paper, we have proposed the novel Feature Separation and Recalibration (FSR) module that recalibrates the non-robust activations to restore discriminative cues that help the model make correct predictions under adversarial attack.
To achieve this goal, FSR first disentangles the intermediate feature map into the robust activations that capture useful cues for correct model decisions and the non-robust activations that are responsible for incorrect predictions.
It then recalibrates the non-robust activations to capture potentially useful cues that could provide additional guidance for more robust predictions on adversarial examples.
We have empirically demonstrated the ability of our method to improve the robustness of various models when applied to different adversarial training strategies across diverse datasets.
We have also verified the superiority of our method to existing approaches that simply deactivate such non-robust activations.

{\small\paragraphTitle{Acknowledgement}{
Prof. Sung-Eui Yoon is a corresponding author.
This work was supported by the National Research Foundation of Korea(NRF) grant funded by the Korea government(MSIT) (No. RS-2023-00208506).
}
}

{\small
\bibliographystyle{ieee_fullname}
\bibliography{egbib}
}

\newpage
\setcounter{table}{0}
\renewcommand{\thetable}{A\arabic{table}}
\setcounter{figure}{0}
\renewcommand{\thefigure}{A\arabic{figure}}

\renewcommand\thesection{\Alph{section}}
\renewcommand\thesubsection{\thesection.\arabic{subsection}}
\addcontentsline{toc}{chapter}{First unnumbered chapter}
\setcounter{section}{0}
\renewcommand*{\theHsection}{chX.\the\value{section}}

\noindent
\textbf{\Large Appendix}

\section{Additional Robustness Evaluation}
In this section, we report the robustness of our FSR on additional datasets (CIFAR-100~\cite{cifar10}, Tiny ImageNet~\cite{learnable}) and model (WideResNet-34-10~\cite{wideresnet}).

\vspace{0.5ex}\noindent
\textbf{Experiments on Other Datasets.}
Table~\ref{table:cifar100} shows the robustness improvements when our FSR module is applied on AT, TRADES, and MART in CIFAR-100 dataset.
While the performance improvements are not as large as in CIFAR-10 and SVHN, applying our FSR module consistently improves the model robustness of all three adversarial training techniques, showing that our method is still effective on more challenging datasets.
We noted that the reason for limited accuracy gain on CIFAR-100 is actually due to its low-resolution data not providing sufficient information for learning the inter-class relationship among cues relevant to various similar classes (e.g., boy and man)~\cite{fine-grained}.

Thus, we also evaluate our method on a more challening Tiny ImageNet dataset with fine-grained classes and higher-resolution images.
As shown by the results in Table~\ref{table:tinyimagenet}, we observed 2.08\% improvement on average for Ensemble robustness compared to vanilla methods, which is significantly higher than that of CIFAR-100 (0.67\%, Table~\ref{table:cifar100}) and on par with CIFAR-10 (2.20\%, Table~\ref{table:robustness-resnet}) and SVHN (2.30\%, Table~\ref{table:robustness-vgg}). 
This shows that our FSR module is also effective on larger, more complex models and datasets and is not limited by the over-parameterization of the model.

\vspace{0.5ex}\noindent
\textbf{Experiments on Other Model.}
In addition to ResNet-18 and VGG16, we also evaluate our FSR module on WideResNet-34-10.
As shown in Table~\ref{table:wideresnet}, our FSR module leads to consistent robustness improvement on WideResNet-34-10.

\section{Additional Ablation Studies}

\vspace{0.5ex}\noindent
\textbf{Position of FSR module.}
Table~\ref{table:position} reports the model robustness when our FSR module is inserted to different layers of ResNet-18. 
As shown in the table, inserting our FSR module after \texttt{Block4} of the model shows the best model robustness under attacks.
This is because the model learns features that are more related to the global semantic information of the image and the final class prediction in the deeper layers, while it learns more low-level features with less semantic information in shallower layers~\cite{cas}.
Recalibrating the non-robust activations in the deeper layers that are more related to the final predictions is more effective at boosting the model robustness.

\begin{table}[t]
	\begin{center}
	    \resizebox{\columnwidth}{!}
		{\begin{tabular}{c|c c c c c c}
        \hline
        \textbf{\textit{ResNet-18}} & \multicolumn{6}{c}{CIFAR-100} \\ \hline 
        Method & Natural & FGSM & PGD-20 & PGD-100 & C\&W & Ensemble \\
        \hline \hline
        {AT}            & \textbf{59.25} & 28.80 & 24.39 & 23.43 & 23.92 & 22.46 \\
        {AT + FSR}      & 58.23 & \textbf{29.58} & \textbf{25.33} & \textbf{24.30} & \textbf{24.54} & \textbf{22.95} \\
        \hline
        {TRADES}        & \textbf{61.87} & 30.77 & 26.37 & 25.76 & 24.08 & 23.45 \\
        {TRADES + FSR}  & 57.27 & \textbf{31.66} & \textbf{27.70} & \textbf{27.27} & \textbf{24.82} & \textbf{24.40} \\
        \hline
        {MART}          & \textbf{57.13} & 31.32 & 27.40 & 26.80 & 25.24 & 24.42 \\
        {MART + FSR}    & 56.51 & \textbf{32.08} & \textbf{27.90} & \textbf{27.28} & \textbf{25.91} & \textbf{24.98} \\
        \hline
        \end{tabular}}
	\end{center}
        \vspace{-3mm}
	\caption{
	    Robustness (accuracy (\%)) of adversarial training strategies (AT, TRADES, MART) with (+ FSR) and without our FSR module against diverse white-box attacks on ResNet-18 and on CIFAR-100 dataset. Better results are marked in \textbf{bold}.
	}
	\label{table:cifar100}
\end{table}

\begin{table}[t]
	\begin{center}
	    \resizebox{\columnwidth}{!}
		{\begin{tabular}{c|c c c c c c}
        \hline
        \textbf{\textit{ResNet-18}} & \multicolumn{6}{c}{Tiny ImageNet} \\ \hline 
        Method & Natural & FGSM & PGD-20 & PGD-100 & C\&W & Ensemble \\
        \hline \hline
        {AT}            & 51.13 & 22.54 & 18.69 & 17.87 & 17.83 & 16.34 \\
        {AT + FSR}      & \textbf{51.77} & \textbf{24.19} & \textbf{20.95} & \textbf{20.06} & \textbf{19.32} & \textbf{18.02} \\
        \hline
        {TRADES}        & \textbf{50.41} & 23.79 & 21.16 & 20.72 & 17.24 & 17.02 \\
        {TRADES + FSR}  & 49.53 & \textbf{24.87} & \textbf{23.22} & \textbf{23.09} & \textbf{19.22} & \textbf{19.04} \\
        \hline
        {MART}          & \textbf{46.21} & 23.84 & 21.75 & 21.35 & 18.34 & 17.71 \\
        {MART + FSR}    & 46.02 & \textbf{26.02} & \textbf{24.05} & \textbf{23.82} & \textbf{20.63} & \textbf{20.24} \\
        \hline
        \end{tabular}}
	\end{center}
        \vspace{-3mm}
	\caption{
	    Robustness (accuracy (\%)) of adversarial training strategies (AT, TRADES, MART) with (+ FSR) and without our FSR module against diverse white-box attacks on ResNet-18 and on Tiny ImageNet dataset. Better results are marked in \textbf{bold}.
	}
	\label{table:tinyimagenet}
\end{table}

\begin{table}[t]
	\begin{center}
	    \resizebox{\columnwidth}{!}
		{\begin{tabular}{c|c c c c c c}
        \hline
        \textbf{\textit{WideResNet-34-10}} & \multicolumn{6}{c}{CIFAR-10} \\ \hline 
        Method & Natural & FGSM & PGD-20 & PGD-100 & C\&W & Ensemble \\
        \hline \hline
        {AT}            & \textbf{87.49} & 59.47 & 50.72 & 48.75 & 50.42 & 48.52 \\
        {AT + FSR}      & 87.02 & \textbf{61.40} & \textbf{53.78} & \textbf{52.04} & \textbf{52.35} & \textbf{50.36} \\
        \hline
        {TRADES}        & 86.06 & 60.78 & 51.77 & 49.66 & 51.34 & 49.27 \\
        {TRADES + FSR}  & \textbf{86.88} & \textbf{62.97} & \textbf{54.37} & \textbf{51.98} & \textbf{53.19} & \textbf{51.34} \\
        \hline
        {MART}          & 85.81 & 61.22 & 52.49 & 49.88 & 49.67 & 48.81 \\
        {MART + FSR}    & \textbf{86.21} & \textbf{62.61} & \textbf{54.23} & \textbf{52.00} & \textbf{51.25} & \textbf{50.10} \\
        \hline
        \end{tabular}}
	\end{center}
        \vspace{-3mm}
	\caption{
	    Robustness (accuracy (\%)) of adversarial training strategies (AT, TRADES, MART) with (+ FSR) and without our FSR module against diverse white-box attacks on WideResNet-34-10 and on CIFAR-10 dataset. Better results are marked in \textbf{bold}.
	}
	\label{table:wideresnet}
\end{table}

\vspace{0.5ex}\noindent
\textbf{Design Choice of $\mathcal{L}_{sep}$.}
As explained in Sec.~\ref{ssec:3.1}, in order to disentangle the non-robust activations through the separation loss $\mathcal{L}_{sep}$ (Eq.~\ref{eq:l_sep}):
\begin{equation}
    \mathcal{L}_{sep} = - \sum^{N}_{i=1} (y_i \cdot \log(p^{+}_i) + y'_i \cdot \log(p^{-}_i)),
\end{equation}
we minimize the cross entropy loss of the prediction score with respect to $y'$, which we define as the label corresponding to the wrong class with the highest prediction score.
In Table~\ref{table:label}, we report the comparison of robustness as we employ different schemes for such disentanglement.
``Uniform" represents replacing $y'$ with a uniform vector implemented through label smoothing,
``Entropy max." represents maximizing the entropy of the output prediction $p^{-}$ on the non-robust feature,
``Avg. targeted loss" represents the average of cross-entropy loss with respect to all class labels except for the ground truth class,
and ``Mispredicted" represents our original design.
All four schemes lead to meaningful improvement compared to the vanilla AT method, as they guide the Separation Net to learn low robustness scores on feature units that are responsible for predictions other than the ground truth class.
Still, our design of using the mispredicted class output achieves the highest robustness under all attacks.
This implies that through this scheme, the Separation Net learns to assign low robustness scores to the most harmful feature units that lead to the most probable model mistake and thus improves the feature robustness by the largest margin.

\begin{table}
	\begin{center}
	    \resizebox{\columnwidth}{!}
		{\begin{tabular}{c|c c c c c c}
        \thickhline
        & No attack & FGSM & PGD-20 & PGD-100 & C\&W & Ensemble \\
        \hline
        \hline
        $\texttt{Block1}$                       & \textbf{84.58}    & 56.41             & 48.29             & 46.28             & 46.96 & 44.89 \\
        $\texttt{Block2}$                       & 83.76             & 56.34             & 48.86             & 47.03             & 47.32 & 45.28 \\
        $\texttt{Block3}$                       & 82.60             & 56.62             & 50.43             & 49.11             & 47.84 & 46.33 \\
        $\texttt{Block4}$                       & 81.46             & \textbf{58.07}    & \textbf{52.47}    & \textbf{51.02}    & \textbf{49.44} & \textbf{48.34} \\
        $\texttt{Block3}$ + $\texttt{Block4}$   & 82.18             & 56.93             & 50.72             & 49.32             & 48.63 & 46.91 
\\
        \thickhline
        \end{tabular}}
	\end{center}
        \vspace{-3mm}
	\caption{
	    Comparison of accuracy (\%) as we insert our FSR module after different layers of ResNet-18.
	}
	\label{table:position}
\end{table}

\begin{table}
	\begin{center}
	    \resizebox{\columnwidth}{!}
		{\begin{tabular}{c|c c c c c c}
        \thickhline
                            & No attack         & FGSM              & PGD-20            & PGD-100           & C\&W      & Ensemble  \\
        \hline
        \hline
        AT                  & 85.02             & 56.21             & 48.22             & 46.37             & 47.38             & 45.51 \\
        \hline
        Uniform             & \textbf{85.16}    & 58.05             & 50.87             & 48.91             & \textbf{49.99}    & 47.90 \\
        Entropy max.        & 84.69             & \textbf{58.35}    & 50.66             & 48.93             & 49.90             & 47.88 \\
        Avg. targeted loss  & 84.50             & 57.98             & 50.41             & 48.55             & 49.80             & 47.42 \\
        Mispredicted (Ours) & 81.46             & 58.07             & \textbf{52.47}    & \textbf{51.02}    & 49.44             & \textbf{48.34} \\
        \thickhline
        \end{tabular}}
	\end{center}
        \vspace{-3mm}
	\caption{
	    Comparison of accuracy (\%) for different design choices of the separation loss $\mathcal{L}_{sep}$ (Eq.~\ref{eq:l_sep}).
	}
	\label{table:label}
\end{table}

\vspace{0.5ex}\noindent
\textbf{Effects of Gumbel Softmax.}
We verify the effects of applying Gumbel softmax to generate a differentiable soft mask $m$ that divides the input feature map into the robust activations and the non-robust activations.
We compare the robustness upon replacing $m$ with a binary mask $b$ (Sec.~\ref{ssec:3.1}) that divides the activations in a discrete manner.
We implement the binary mask $b$ by first applying a sigmoid normalization function to the robustness map $r$ generated by the Separation Net and setting all values less than 0.5 to 0 and all values greater than or equal to 0.5 to 1.
In other words, for an $i$-th unit of the robustness map $r$, we set $b_i$ as follows:
\begin{equation}
    b_i = \begin{cases}
        0, & \text{if}\ \sigma(r)_i < t \\
        1, & \text{if}\ \sigma(r)_i \ge t,
    \end{cases}
\end{equation}
where $t = 0.5$, and $\sigma(\cdot)$ is the sigmoid normalization function.

In Table~\ref{table:gumbel}, we show the comparison of robustness of our method upon using either $b$ (Binary) or $m$ (Gumbel).
Using the differentiable mask $m$ through the Gumbel softmax leads to higher robustness against all white-box attacks and especially against the AutoAttack than using the binary mask $b$.
Using the Gumbel softmax allows us to learn the mask to better capture the feature robustness, and it also prevents gradient masking, thus showing higher robustness against AutoAttack.

\begin{table}
	\begin{center}
	    \resizebox{\columnwidth}{!}
		{\begin{tabular}{c|c c c c c c}
        \thickhline
                & FGSM              & PGD-20            & PGD-100           & C\&W              & Ensemble  & AutoAttack  \\
        \hline
        \hline
        Binary  & 55.78             & 49.21             & 47.79             & 48.74             & 46.91     & 44.26 \\
        Gumbel  & \textbf{58.07}    & \textbf{52.47}    & \textbf{51.02}    & \textbf{49.44}    & \textbf{48.34}    & \textbf{46.41} \\
        \thickhline
        \end{tabular}}
	\end{center}
        \vspace{-3mm}
	\caption{
	    Comparison of accuracy (\%) on using mask generated by discrete binary sampling or through Gumbel softmax.
	}
	\label{table:gumbel}
\end{table}

\begin{table}[t]
	\begin{center}
	    \resizebox{\columnwidth}{!}
		{\begin{tabular}{c|c c c c c}
        \thickhline
        & FGSM & PGD-20 & PGD-100 & C\&W & Ensemble \\
        \hline
        \hline
        Greedy          & 57.75             & 49.48             & 47.59             & 48.36             & 46.42 \\
        Random          & 56.60             & 50.04             & 48.46             & 49.08             & 46.77 \\
        w/o Separation        & 57.51             & 50.71             & 48.98             & 49.32             & 47.60 \\
        w/ Separation (Ours)  & \textbf{58.07}    & \textbf{52.47}    & \textbf{51.02}    & \textbf{49.44}    & \textbf{48.34} \\
        \thickhline
        \end{tabular}}
	\end{center}
        \vspace{-3mm}
	\caption{
	    Comparison of accuracy (\%) as we replace the Separation Net with different strategies.
	}
	\label{table:separation}
\end{table}

\vspace{0.5ex}\noindent
\textbf{Experiments on Effectiveness of the Separation Net.}
In order to verify whether our Separation Net is learning appropriate robustness scores for each feature activaiton, we tried replacing the output mask $m$ from the Separation Net (Eq.~\ref{eq:mask}) with different strategies.
We tested random selection and a greedy method of recalibrating the lowest activations, both of which would recalibrate feature activations unaware of their robustness. 
Table~\ref{table:separation} shows that both strategies significantly lag behind our method without Separation, which is equivalent to recalibrating all activations (refer to Table~\ref{table:abl-sep}). 
This is because they do not fully recapture the discriminative cues underlying in non-robust activations. 
Our method with Separation leads to the highest robustness, showing that FSR well identifies the non-robust activations and recaptures discriminative cues from them.

\begin{figure*}[t]
    \centering
    \begin{subfigure}[t]{0.33\textwidth}
        \centering
        \includegraphics[width=\textwidth]{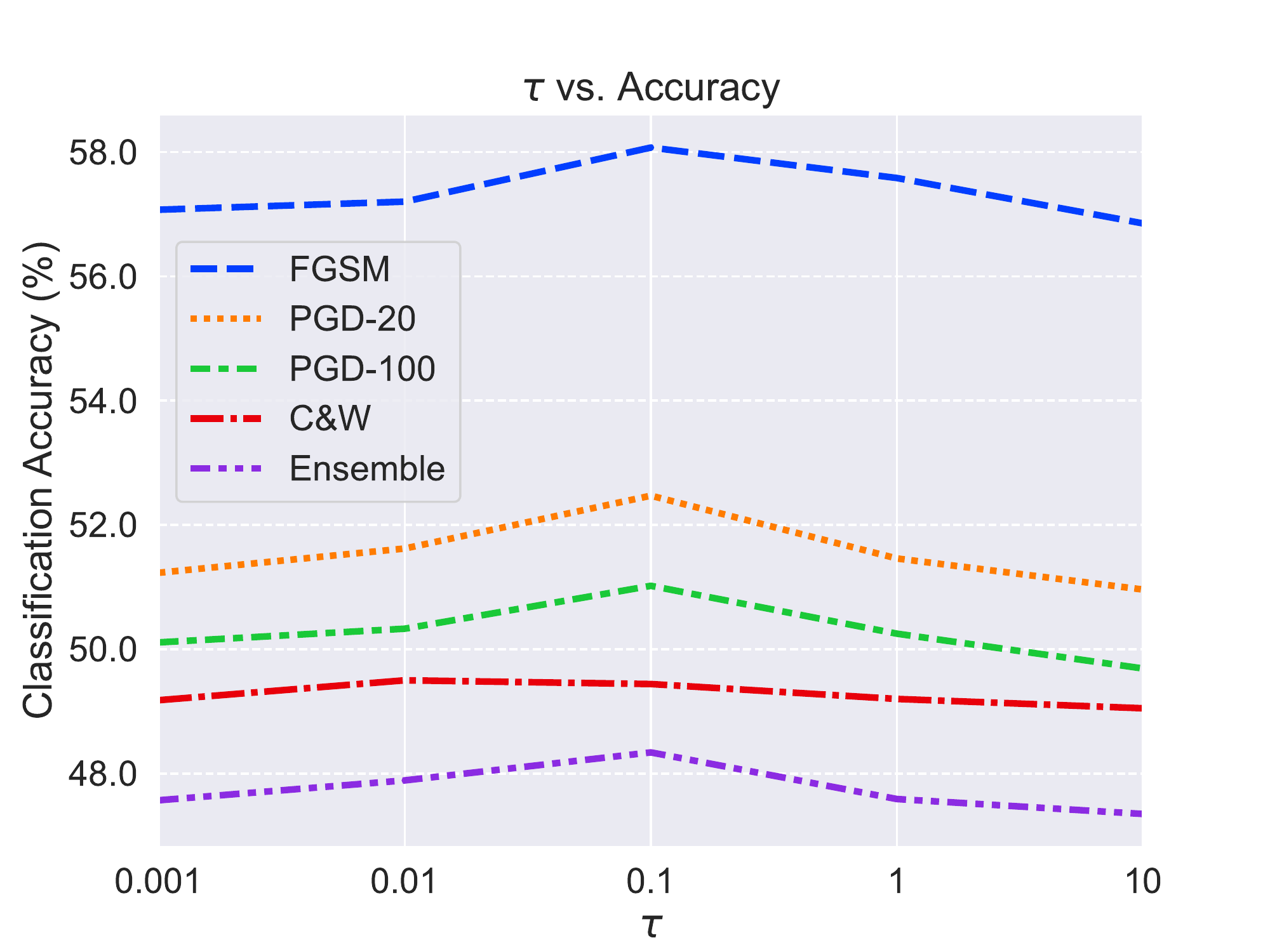}
        \caption{$\tau$}
        \label{fig:tau}
    \end{subfigure}%
    \hfill
    \begin{subfigure}[t]{0.33\textwidth}
        \centering
        \includegraphics[width=\textwidth]{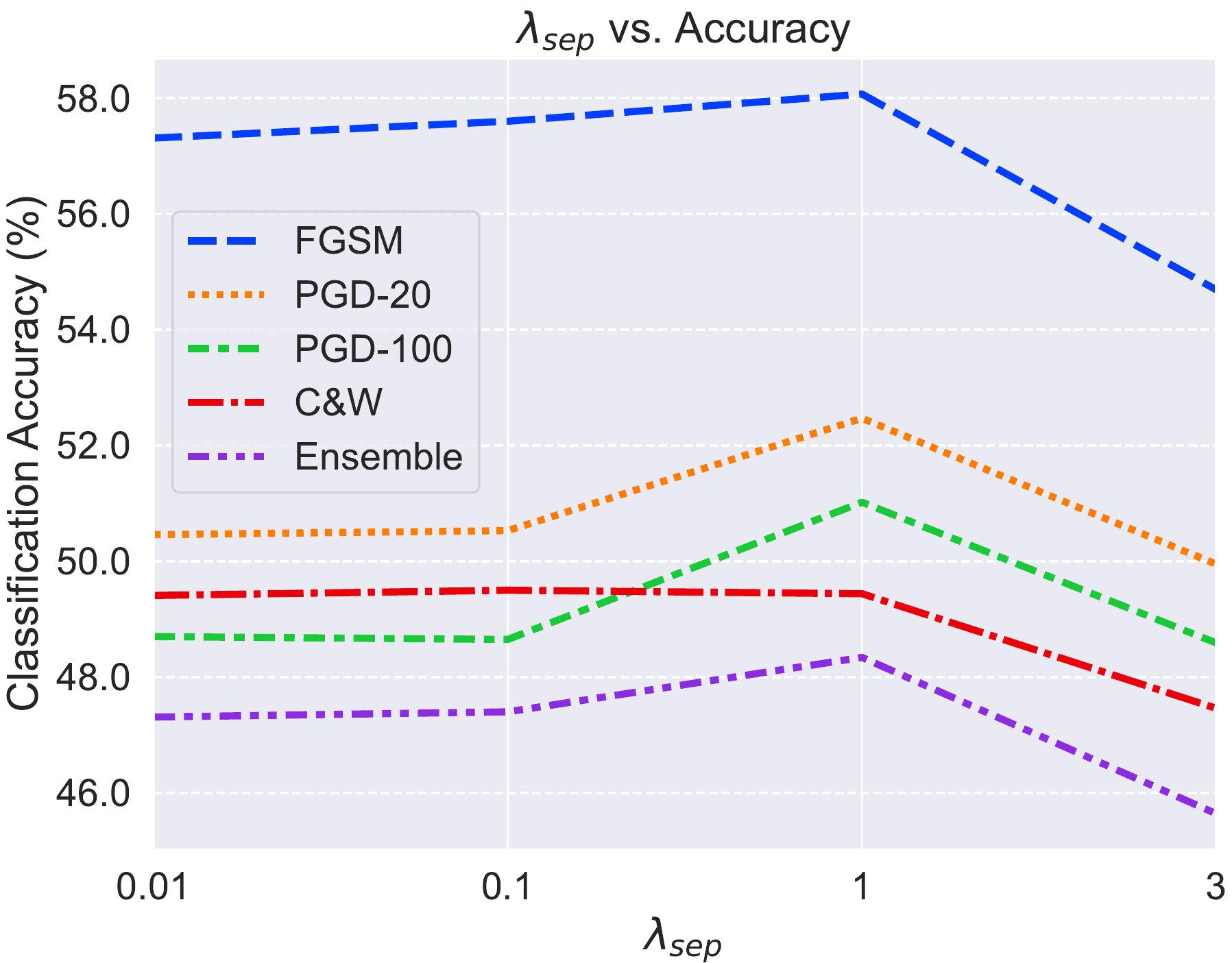}
        \caption{$\lambda_{sep}$}
        \label{fig:lam-sep}
    \end{subfigure}%
    \hfill
    \begin{subfigure}[t]{0.33\textwidth}
        \centering
        \includegraphics[width=\textwidth]{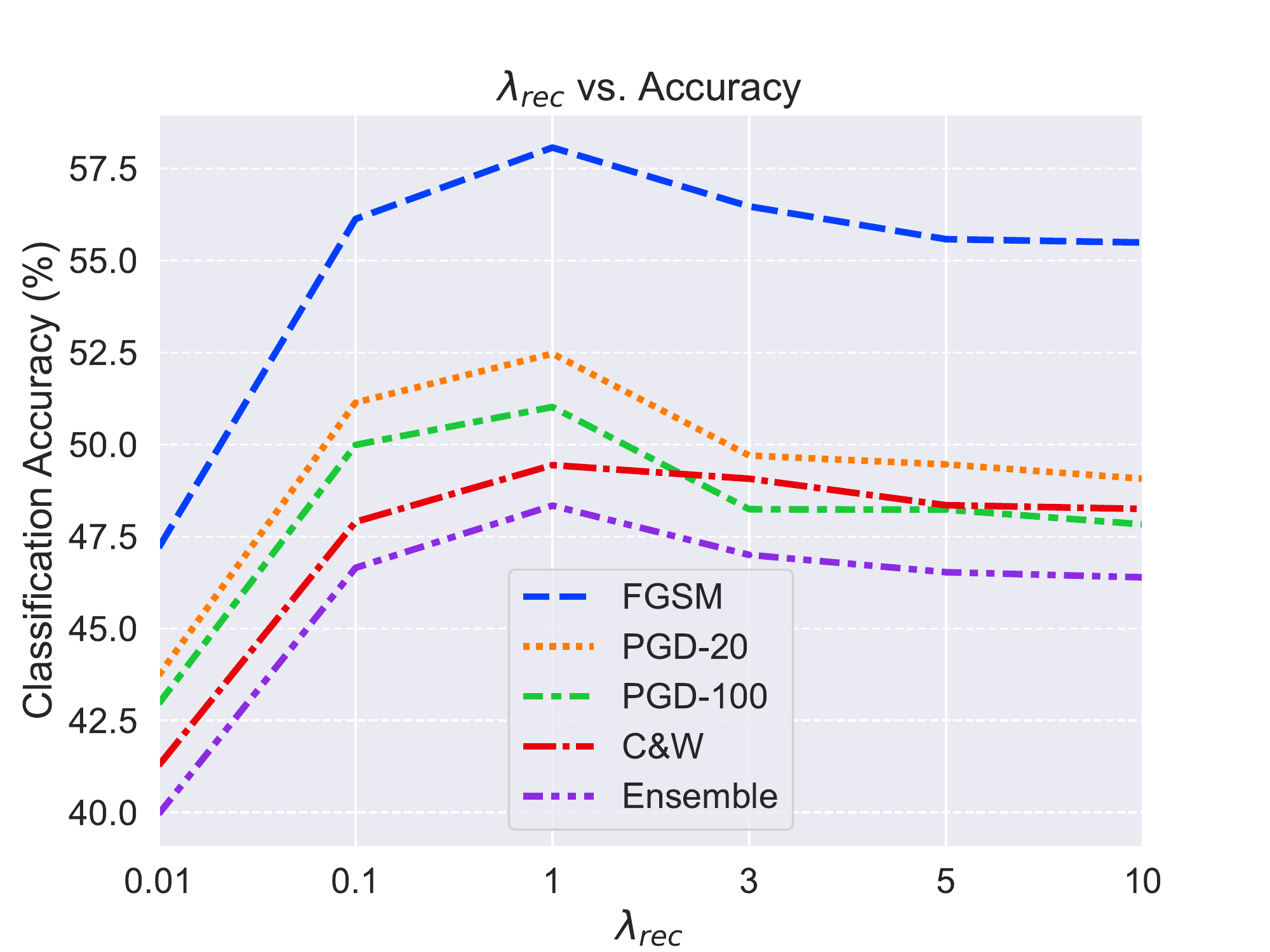}
        \caption{$\lambda_{rec}$}
        \label{fig:lam-rec}
    \end{subfigure}
    \caption{
        Analysis on the robustness with various values of hyperparameters used in FSR module.
        (a) Study on $\tau$ that controls the temperature on Gumbel softmax.
        (b) Study on $\lambda_{sep}$ that controls the weight on the separation loss $\mathcal{L}_{sep}$.
        (c) Study on $\lambda_{rec}$ that controls the weight on the recalibration loss $\mathcal{L}_{rec}$.
    }
    \label{fig:hyperparameter}    
\end{figure*}

\begin{figure}[t]
    \centering
    \begin{subfigure}[t]{0.5\columnwidth}
        \centering
        \includegraphics[width=\textwidth]{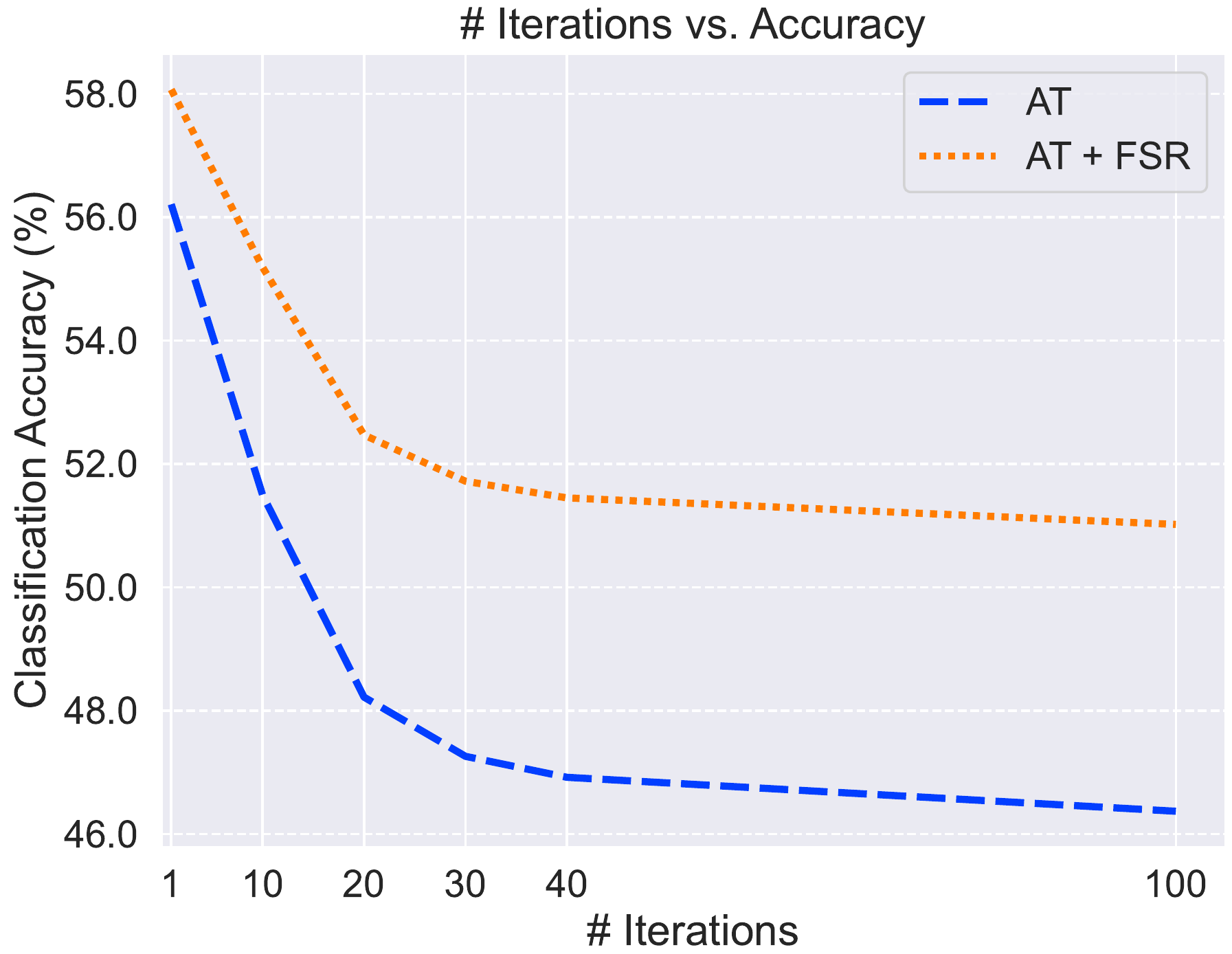}
        \caption{Varying \# iterations}
        \label{fig:iter}
    \end{subfigure}%
    \hfill
    \begin{subfigure}[t]{0.5\columnwidth}
        \centering
        \includegraphics[width=\textwidth]{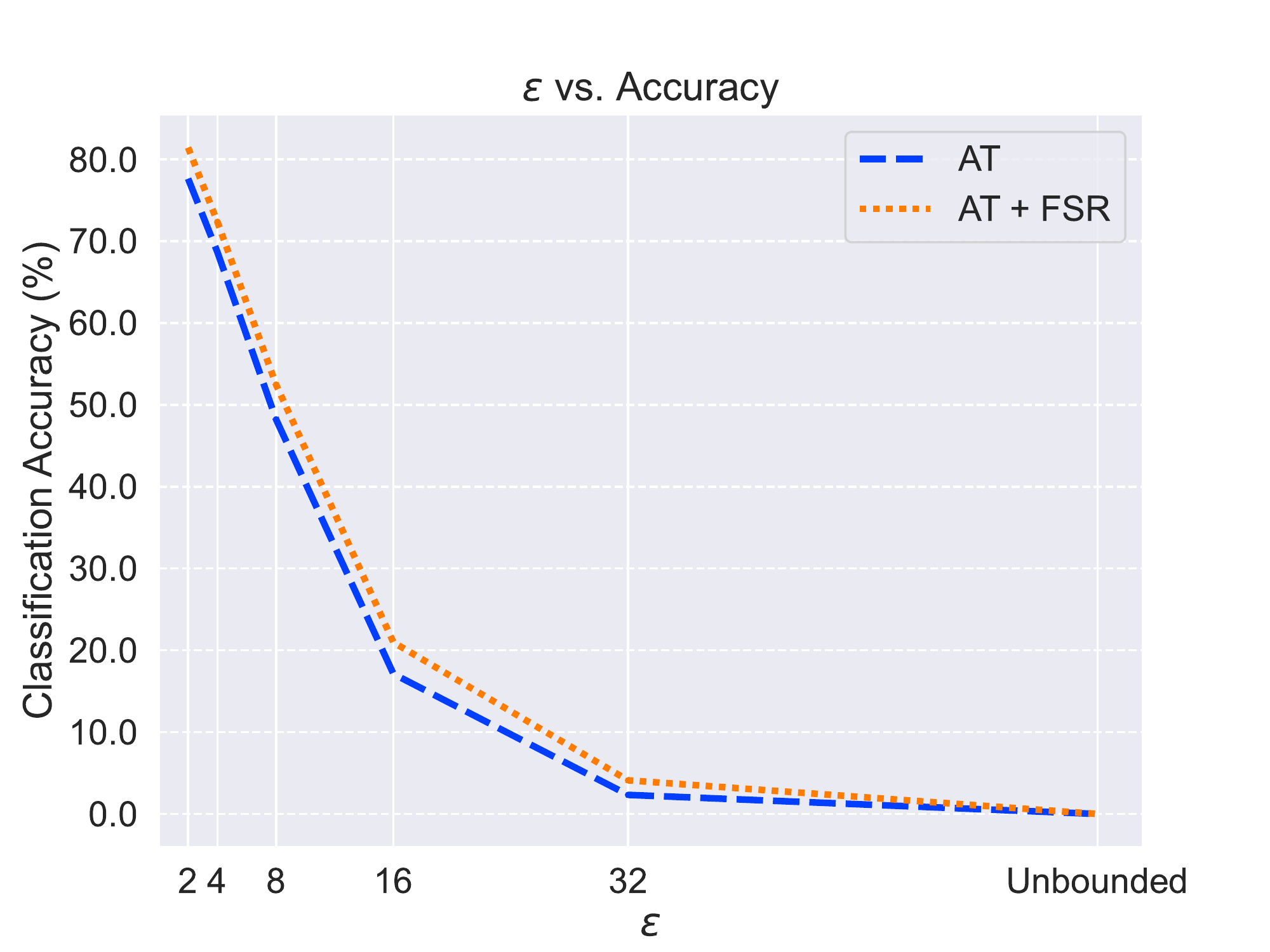}
        \caption{Varying $\epsilon$}
        \label{fig:eps}
    \end{subfigure}
    \caption{
        Analysis on obfuscated gradients.
        (a) Change in model robustness as we vary the number of iterations in PGD attack.
        (b) Change in model robustness as we vary the perturbation bound $\epsilon$.
    }
    \label{fig:obfuscated}    
\end{figure}

\vspace{0.5ex}\noindent
\textbf{Hyperparameter Study.}
We also compare the robustness as we vary the temperature $\tau$ (Eq.~\ref{eq:mask}) that controls how ``discrete" the mask is.
For low temperature values, the output mask becomes more discrete (\ie, most values are close to either 0 or 1), and for high temperature values, it becomes more uniform (\ie, most values are far away from 0 or 1)~\cite{gumbel}.
As shown in Fig.~\ref{fig:tau}, we achieve the highest robustness when $\tau = 0.1$.
From this observation, we can see that too small $\tau$ will degenerate the Gumbel softmax into binary sampling and make the mask become a binary mask, which could result in no gradients or improper training~\cite{spatial}.
In contrast, too large $\tau$ will make the mask become more uniformly distributed and reduce the gap between the mask values applied on robust or non-robust activations, thus making our goal of disentanglement less feasible.

In Fig.~\ref{fig:lam-sep} and Fig.~\ref{fig:lam-rec}, we visualize the trends of model robustness as we vary the weights on our proposed loss functions $\mathcal{L}_{sep}$ (Eq.~\ref{eq:l_sep}) and $\mathcal{L}_{rec}$ (Eq.~\ref{eq:l_rec}).
Higher value of $\lambda_{sep}$ generally improves robustness under all attacks with the best performance achieved when $\lambda_{sep} = 1$, showing that our proposed objectives help the model learn more robust feature representations.
Similar trends can also be observed for $\lambda_{rec}$; higher value of $\lambda_{rec}$ generally improves robustness with the best performance achieved when $\lambda_{rec} = 1$.
Setting $\lambda_{sep}$ and $\lambda_{rec}$ to be too high, however, tends to degrade robustness.
This is because of the trade-off between the vanilla classification loss $\mathcal{L}_{cls}$ on the final classifier layer and the two auxiliary loss.
As we focus more on the objectives on the auxiliary layer, the two auxiliary losses may deviate the model from learning the classification task based on $\mathcal{L}_{cls}$.

\section{Analysis on Obfuscated Gradients}
In this section, we verify that the robustness of our method is not a result of obfuscating gradients.
We test our method under the following criteria~\cite{obfuscated} to demonstrate that our method does not obfuscate gradients:
\begin{enumerate}[label=(\roman*)]
    \itemsep0em 
    \item\label{item:first} White-box attacks are stronger than black-box attacks,
    \item\label{item:second} Robustness decreases with the increased number of iterations in gradient-based attacks,
    \item\label{item:third} Robustness decreases with increased perturbation bound $\epsilon$, and unbounded attacks achieve 100\% attack success rate.
\end{enumerate}

Tables~\ref{table:robustness-resnet} and~\ref{table:black-box} show the robustness of our method under both white-box and black-box attacks when applied to ResNet-18 on the CIFAR-10 dataset.
Comparing the two tables, we can observe that the strongest black-box attacks (\eg, DI-FGSM and $\mathcal{N}$Attack) are still weaker than white-box attacks (\eg, C\&W), meeting the requirement~\ref{item:first}.
Fig.~\ref{fig:iter} shows robustness of our method and vanilla PGD adversarial training under PGD attacks with various number of iterations. 
The robustness does indeed decrease with increasing number of iterations, meeting the requirement~\ref{item:second}.
Fig.~\ref{fig:eps} shows robustness of the two methods under PGD attacks with various perturbation bounds $\epsilon$ under $\ell_\infty$-norm.
Similarly, the robustness decreases with increasing $\epsilon$, and it reaches 0\% accuracy under unbounded attacks, thus meeting the requirement~\ref{item:third}.

\end{document}